%% file: main.tex
\documentclass{article}
\pdfoutput=1  

\usepackage{hyperref}

\usepackage[accepted]{icml2024}  

\input{preamble}
\usepackage{listings}

\input{math_commands}

\input{symbols}

\icmltitlerunning{Decision Mamba: Reinforcement Learning via Sequence Modeling with Selective State Spaces}

\begin{document}
\twocolumn[
\icmltitle{Decision Mamba: Reinforcement Learning via Sequence Modeling \\ with Selective State Spaces}

\icmlsetsymbol{equal}{*}
\begin{icmlauthorlist}
\icmlauthor{Toshihiro Ota}{ca,riken}
\end{icmlauthorlist}

\icmlaffiliation{ca}{AI Lab, CyberAgent, Japan}
\icmlaffiliation{riken}{RIKEN iTHEMS, Japan}
\icmlcorrespondingauthor{Toshihiro Ota}{ota\_toshihiro@cyberagent.co.jp}

\icmlkeywords{Machine Learning, Reinforcement Learning, State Space Models}
\vskip 0.3in
]

\printAffiliationsAndNotice{}  

\input{abstract}
\input{sec1}
\input{sec2}
\input{sec3}
\input{sec4}
\input{sec5}

\newpage
\bibliography{references}
\bibliographystyle{icml2024}

\newpage
\appendix
\onecolumn
\input{appendix}

\end{document}

%% file: preamble.tex
\usepackage{amsmath}
\usepackage{amssymb}
\usepackage{mathtools}
\usepackage{amsthm}

\usepackage[capitalize,noabbrev]{cleveref}

\theoremstyle{plain}

\theoremstyle{definition}

\theoremstyle{remark}

\usepackage[textsize=tiny]{todonotes}

\usepackage[utf8]{inputenc} 
\usepackage[T1]{fontenc}    
\usepackage{url}            
\usepackage{booktabs}       
\usepackage{amsfonts}       
\usepackage{nicefrac}       
\usepackage{microtype}      
\usepackage{xcolor}         

\usepackage{bm}
\usepackage{graphicx} 
\usepackage{caption} 
\usepackage[subrefformat=parens]{subcaption}
\usepackage{comment}

%% file: math_commands.tex
\usepackage{amsmath,amsfonts,bm}

\def\eqref#1{equation~\ref{#1}}

\def\1{\bm{1}}

\def\eps{{\epsilon}}

\def\rb{{\textnormal{b}}}

\def\rh{{\textnormal{h}}}

\def\rt{{\textnormal{t}}}

\def\vb{{\bm{b}}}

\def\vh{{\bm{h}}}

\def\vt{{\bm{t}}}

\DeclareMathAlphabet{\mathsfit}{\encodingdefault}{\sfdefault}{m}{sl}
\SetMathAlphabet{\mathsfit}{bold}{\encodingdefault}{\sfdefault}{bx}{n}

\newcommand{\E}{\mathbb{E}}

\newcommand{\R}{\mathbb{R}}

\let\ab\allowbreak

%% file: symbols.tex
\newcount\K 
\def\blah{
    \K=0 \loop\ifnum\K<20 
    {\color[rgb]{0.8, 0.8, 0.8} blah blah blah blah blah blah blah blah blah blah} 
    \advance\K by1\repeat 
}

\usepackage{mleftright}
\mleftright

\newcommand{\mdp}{\mathcal{M}}
\newcommand{\sspace}{\mathcal{S}}
\newcommand{\aspace}{\mathcal{A}}
\newcommand{\rewardfun}{\mathcal{R}}
\newcommand{\policy}{\pi}
\newcommand{\rtg}{R}  
\newcommand{\traj}{\tilde{\tau}}

\newcommand{\Amat}{\bm{A}}
\newcommand{\Bmat}{\bm{B}}
\newcommand{\Cmat}{\bm{C}}
\newcommand{\Abar}{\overline{\bm{A}}}
\newcommand{\Bbar}{\overline{\bm{B}}}

\DeclareMathOperator{\LN}{LayerNorm}
\DeclareMathOperator{\gelu}{GELU}
\DeclareMathOperator{\silu}{SiLU}
\DeclareMathOperator{\linear}{Linear}

\newcommand{\model}{DMamba}
\newcommand{\block}{MambaBlock}
\newcommand{\githubrepo}{https://github.com/Toshihiro-Ota/decision-mamba}

\newcommand{\eref}[1]{Eq.~(\ref{#1})}
\newcommand{\fref}[1]{Fig.~\ref{#1}}
\newcommand{\tref}[1]{Table~\ref{#1}}

\newcommand{\aref}[1]{Appendix~\ref{#1}}

\newcommand{\set}[1]{\left\{ #1 \right\}}

%% file: abstract.tex
\begin{abstract}

Decision Transformer, a promising approach that applies Transformer architectures to reinforcement learning, relies on causal self-attention to model sequences of states, actions, and rewards.
While this method has shown competitive results, this paper investigates the integration of the Mamba framework, known for its advanced capabilities in efficient and effective sequence modeling, into the Decision Transformer architecture, focusing on the potential performance enhancements in sequential decision-making tasks.
Our study systematically evaluates this integration by conducting a series of experiments across various decision-making environments, comparing the modified Decision Transformer, \emph{Decision Mamba}, with its traditional counterpart.
This work contributes to the advancement of sequential decision-making models, suggesting that the architecture and training methodology of neural networks can significantly impact their performance in complex tasks, and highlighting the potential of Mamba as a valuable tool for improving the efficacy of Transformer-based models in reinforcement learning scenarios.

\end{abstract}

%% file: sec1.tex
\section{Introduction}

The interplay between sequence-based decision-making and large-scale model efficiency presents a fascinating subject in the field of modern machine learning.
Decision Transformer \cite{chen2021decision} represents a paradigm shift in reinforcement learning, introducing a sequence-to-sequence modeling that replaces traditional reinforcement learning's reliance on value functions with a direct mapping of state-action-reward sequences to optimal actions using causal self-attention mechanisms.
While the use of causal self-attention has been instrumental in achieving remarkable results, the exploration of alternative mechanisms that could further elevate model performance remains an open and compelling question.

Recently, the Mamba framework \cite{gu2023mamba} has been proposed as an efficient and effective sequence modeling framework endowed with the selective structured state space model.
Mamba introduces the data-dependent selection mechanism with efficient hardware-aware design to tackle the data- and time-invariant issues of prior state space models \cite{gu2020hippo,gu2021combining,gu2022efficiently}.
This allows the model to selectively extract essential information and filter out irrelevant noise according to the input data, leading to superior sequential modeling performance.
Owing to these innovative approaches, it has emerged as a pivotal solution for the efficient training of massive neural networks, optimizing parallel processing of data and models across multiple computing units.

This paper sets out to explore the integration of the Mamba framework as a novel architectural choice within Decision Transformer, with a focus on the potential performance improvements this integration could bring.
We introduce \emph{Decision Mamba} by substituting Mamba for causal self-attention, aiming to investigate the extent to which this modification can enhance the model's ability to capture complex dependencies and nuances in sequential decision-making tasks, potentially leading to superior decision-making capabilities in a variety of challenging environments.
While efficiency is a notable benefit, our primary focus in this paper lies in the possible performance gains that Mamba's architecture design might confer on the Decision Transformer's sequence modeling capabilities.
Specifically, we hypothesize that Mamba's design could offer a novel way to encode and exploit the temporal dependencies and intricate patterns present in sequential decision-making tasks, potentially resulting in more accurate, robust, and nuanced decision-making outputs.

Through this investigation, we seek to not only expand the tools available for reinforcement learning challenges but also to provide insights into the interplay between different architectural components and their impact on model performance in complex decision-making environments.
By examining the performance implications of integrating Mamba into Decision Transformer, this work aims to contribute to the broader discourse on how best to architect and train models for the nuanced demands of sequential decision-making, offering potential pathways to significant advancements in the field.

%% file: sec2.tex
\section{Preliminaries}

\subsection{Offline reinforcement learning}

In a conventional reinforcement learning (RL) problem, an environment is modeled as a Markov decision process (MDP) $\mdp=(\sspace, \aspace, p_0, P, \rewardfun)$,
which consists of states $s\in \sspace$, actions $a\in \aspace$, the initial state distribution $p_0$, the transition probability function $P(s',s,a)$, and the reward function $r=\rewardfun(s,a)$.
The interaction between an agent and the environment is represented as a trajectory that is a sequence of states, actions, and rewards for each timestep: $\tau = (s_0, a_0, r_0, s_1, a_1, r_1, \ldots, s_T, a_T, r_T)$.
The goal of conventional RL is to learn a policy $\policy$ that maximizes the expected return $\E_{\mdp, \policy} \left[ \sum_{i=0}^T r_i \right]$.
In offline RL, or batch RL, we only have access to a fixed amount of a previously collected dataset and learning is performed without interaction with the environment, which is known to be much harder than online settings \cite{levine2020offline}.

An approach we can utilize is Behavior Cloning (BC) \cite{bain1995framework}, which directly learns the mapping from state to action from the dataset.
The offline RL dataset, however, often lacks sufficient expert demonstrations.
To resolve this issue, we may adopt return-conditioned BC.
This method uses reward information and takes a target future return as input from the dataset.
Namely, we define the return of a trajectory at timestep $i$, $\rtg_i=\sum_{i'=i}^{T} r_{i'}$, which is referred to as the return-to-go (RTG),
and feed a model with the following return-conditioned trajectory for autoregressive learning:
\begin{equation}
    \traj = (\rtg_0, s_0, a_0, \rtg_1, s_1, a_1, \ldots, \rtg_T, s_T, a_T).
    \label{eq:traj}
\end{equation}

Decision Transformer (DT) \cite{chen2021decision} not only provides a promising approach to return-conditioned BC but also reconceptualizes the RL problems as a sequence modeling task.
DT essentially operates as a causal transformer model, taking a sequence of RTGs, states, and actions as inputs, as in \eref{eq:traj}.
DT autoregressively learns the trajectories for predicting the optimal action given the target RTG.
Despite not using traditional RL tools such as the value function or the Bellman operator, DT has demonstrated its effectiveness in empirical studies.

\subsection{State Space Models}

The State Space Models (SSMs) in deep learning are a class of sequence modeling frameworks based on linear ordinary differential equations.
They map an input signal $x(t) \in \R^D$ to the output signal $y(t) \in \R^D$ via the latent state $h(t) \in \R^N$:
\begin{align}
    h'(t) &= \Amat h(t) + \Bmat x(t),  \label{eq:ssm}  \\
    y(t)  &= \Cmat h(t),
\end{align}
where $\Amat \in \R^{N \times N}$ and $\Bmat, \Cmat^{\top} \in \R^{N \times D}$ will be trainable matrices.
For application to a discrete input sequence $(x_0, x_1, \ldots)$ instead of a continuous function, \eref{eq:ssm} is discretized with a step size $\Delta$, indicating the input's resolution.
Various discretization rules can be used such as the Euler method or the so-called Bilinear.
We will be considering the discretized SSM using the zero-order hold (ZOH) discretization rule:
\begin{align}
    h_t &= \Abar h_{t-1} + \Bbar x_t,  \\
    y_t &= \Cmat h_t,
\end{align}
where $\Abar=\exp (\Delta \Amat)$ and $\Bbar = (\Delta \Amat)^{-1}(\exp (\Delta \Amat) - \bm{I}) (\Delta \Bmat)$.
By converting the continuous SSM to the discrete one and transforming the parameters from $(\Delta, \Amat, \Bmat)$ to $(\Abar, \Bbar)$, the model becomes a sequence-to-sequence mapping framework from $\set{x_t}$ to $\set{y_t}$.
Imposing structural conditions on the state matrix $\Amat$ with the HiPPO initialization \cite{gu2020hippo}, this expression gives rise to the structured state space model (S4) \cite{gu2022efficiently},
broadly related to recurrent neural networks and convolutional neural networks \cite{gu2021combining,gupta2022diagonal,gu2022parameterization}.

Based on the S4 framework, Mamba \cite{gu2023mamba} introduces a data-dependent selection mechanism while leveraging a hardware-aware parallel algorithm in recurrent mode.
The combined architecture of Mamba empowers it to capture contextual information effectively, especially for long sequences, and maintains computational efficiency.
Being a linear-time sequence model, Mamba delivers Transformer-quality performance with improved efficiency, particularly for long sequences.

%% file: sec3.tex
\section{Decision Mamba}

We introduce Decision Mamba (\model) by utilizing the Mamba block as a token-mixing module instead of the self-attention module of DT.
In this section, we begin with an overview of our architecture and describe \model ~as a variant of the Transformer-type networks.

\begin{figure}[t] 
    \centering
    \includegraphics[keepaspectratio, scale=0.36]{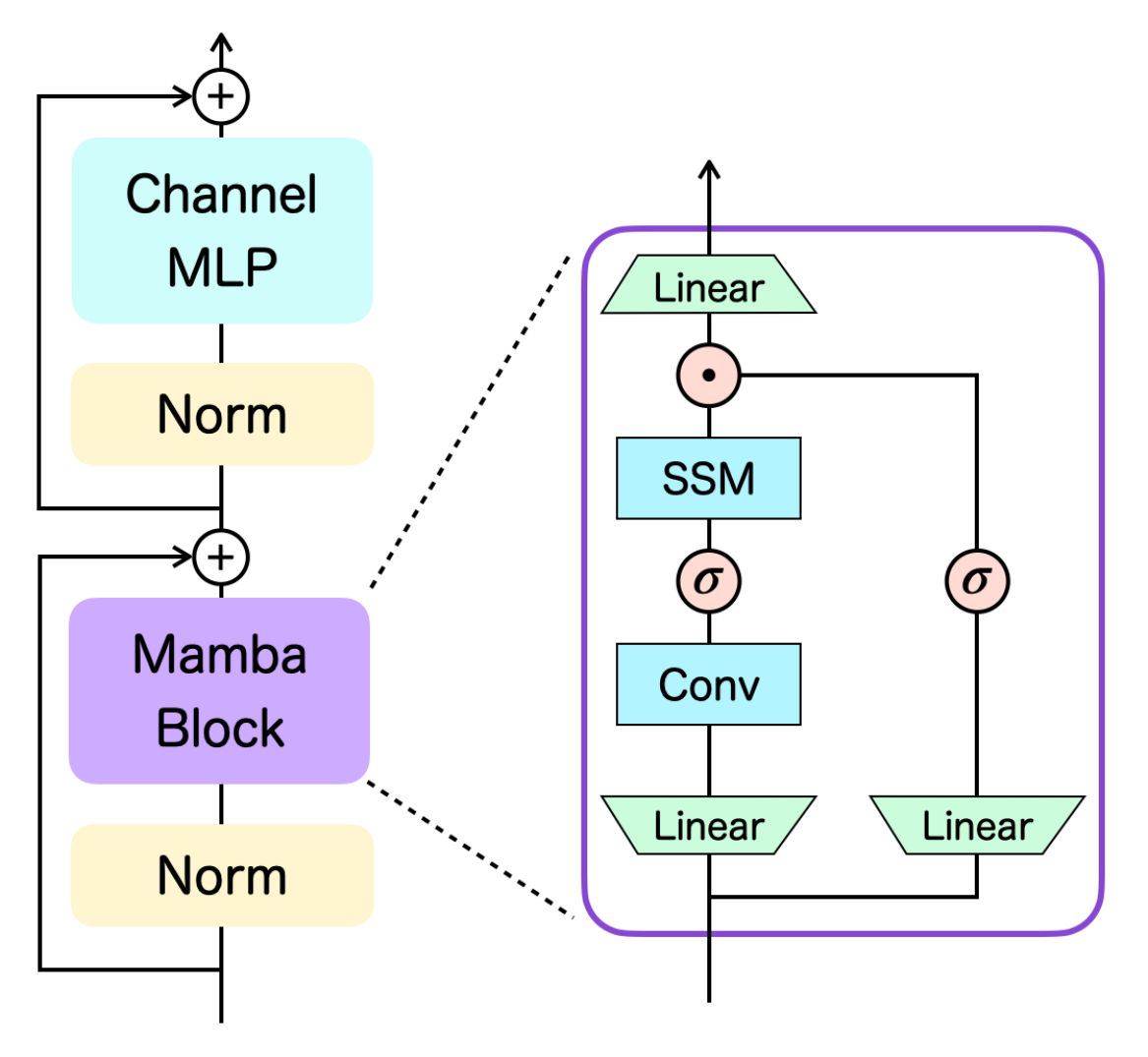}
    \caption{Overview of the Mamba layer.
            $\sigma$ is an activation function, for which we use the $\silu$ function,
            and $\odot$ denotes the element-wise product.}
    \label{fig:mambablock}
\end{figure}

\textbf{Network architecture.}
The network architecture of \model ~adopts the basic Transformer-type neural network, essentially the GPT architectures \cite{radford2018improving}.
The main module, the Mamba layer, is composed of the token-mixing layer and the channel-mixing layer,
both of which consist of the layer normalization, the residual connection, and the corresponding mixing block, see the left-hand side of \fref{fig:mambablock}.
We feed the last $K$ timesteps of a trajectory into \model, for the total of $3K$ tokens:
\begin{equation}
    \traj_i = (\rtg_{i-K+1}, s_{i-K+1}, a_{i-K+1}, \ldots, \rtg_i, s_i, a_i).
\end{equation}
The input trajectory $\traj_i$ is transformed into token embeddings $I_i$ by an embedding layer, which is either a linear layer or a two-dimensional convolutional layer depending on the environments,
\begin{equation}
    I_i = \operatorname{Emb}(\traj_i) \in \R^{3K \times D}.
\end{equation}
Then the input sequence of tokens is processed by a stack of Mamba layers,
\begin{align}
    U^l     &= X^l + \operatorname{\block}\left( \LN(X^l) \right),  \\
    X^{l+1} &= U^l + \operatorname{ChannelMLP}\left( \LN(U^l) \right),  \label{eq:channelmlp}
\end{align}
where $l$ runs from $0$ to the total number of layers and $X^0 = I_i$.
The detail of the Mamba block is described below.
The rest of the network is identical to that of DT.
To make the comparisons fair in the subsequent section, our Mamba layer involves the channel-mixing MLP in \eref{eq:channelmlp}, also known as the point-wise feedforward network,
though it is not necessarily included in the original Mamba model in \cite{gu2023mamba}.
The pseudo-code of the Mamba layer is provided in \aref{appendix:code}.

\textbf{Mamba block.}
The right-hand side of \fref{fig:mambablock} and Algorithm~\ref{alg:mamba} illustrate the series of operations performed inside the Mamba block.
Let us suppose that the input sequence of tokens $x$ has shape $(B, L, D)$, where $B$ is the batch size,
$L$ represents the sequence length corresponding to both the indices $t$ of the discretized SSM and to $3K$ of the trajectory length, and $D$ is the embedding dimension of the channels.
The Mamba block first operates linear projections to obtain hidden states $x$ and $z$, with a controllable expansion factor $E$.
Activated by the $\silu$ function after the causal one-dimensional convolution along the sequence length,
it transforms the hidden state $x$ to make $\Bmat$, $\Cmat$ and $\Delta$ be functions of the input:
\begin{align}
    \Bmat  &= \linear(x),  \quad  \Cmat = \linear(x),  \\
    \Delta &= \operatorname{Softplus}\left( \mathrm{Parameter} + s_\Delta(x) \right),
\end{align}
where $s_\Delta(x) = \operatorname{Broadcast}(\linear(x))$.
The ZOH discretization gives $\Abar$ and $\Bbar$, which generate the core of the block, the selective SSM.
Finally, the product with the hidden state $z$ applied by another $\linear$ and $\silu$ functions delivers the output $y$.
In this block, we have several controllable hyperparameters including $N$, $E$, and the kernel size of one-dimensional convolution.
Instead of searching the best hyperparameters for \model, in this paper we employ the default ones of the original implementation,%
\footnote{\url{https://github.com/state-spaces/mamba}}
$N=16$ and $E=2$, etc.

\begin{algorithm}[t] 
    \caption{Mamba block with selective SSM.}
    \label{alg:mamba}
    \begin{algorithmic}[1]
        \renewcommand{\algorithmicrequire}{\textbf{Input:}}
        \renewcommand{\algorithmicensure}{\textbf{Output:}}
        \REQUIRE $x: (B, L, D)$
        \ENSURE $y: (B, L, D)$

        \STATE $x,z: (B, L, ED)  \leftarrow  \linear(x)$
        \STATE $x: (B, L, ED)  \leftarrow  \silu(\operatorname{Conv1d}(x))$

        \STATE $\Amat: (D, N)  \leftarrow  \mathrm{Parameter}$
        \STATE $\Bmat, \Cmat: (B, L, N)  \leftarrow  \linear(x), \linear(x)$
        \STATE $\Delta: (B, L, D)  \leftarrow  \operatorname{Softplus}(\mathrm{Parameter} + s_\Delta(x))$
        \STATE $\Abar,\Bbar: (B, L, D, N)  \leftarrow  \operatorname{ZOH}(\Delta, \Amat, \Bmat)$

        \STATE $y: (B, L, ED)  \leftarrow  \operatorname{SelectiveSSM}(\Abar, \Bbar, \Cmat)(x)$
        \STATE $y: (B, L, ED)  \leftarrow  y \odot \silu(z)$
        \STATE $y: (B, L, D)  \leftarrow  \linear(y)$

        \STATE \textbf{return} $y$
    \end{algorithmic}
\end{algorithm}

\textbf{Training \& inference.}
In the training phase, we have a dataset of offline trajectories.
We sample sub-trajectories of length $3K$, $\traj$, from the dataset.
Given the current state $s_i$ and the RTG $\rtg_{i}$, the model predicts the next action $\hat{a}_i$.
The objective function to optimize depends on the environments: the mean squared error (MSE) for continuous actions or the cross-entropy loss (CE) for discrete actions.
The losses for each timestep are then averaged:
\begin{equation}
    \mathrm{Loss} = \E_{\traj \sim \mdp,\policy} \left[ \frac1K \sum_{i=0}^{K-1} \mathcal{L}_{\text{MSE/CE}}(\hat{a}_i; a_i) \right].
\end{equation}
In the inference phase, we set a target RTG $\rtg_0$, representing the desired performance, as the initial condition.
During the inference, the model receives a current trajectory and generates an action.
This results in obtaining a next state and a reward $r_i$.
Subsequently, it subtracts this reward from the RTG of the previous timestep, $\rtg_{i-1}$.

%% file: sec4.tex
\section{Experiments}

In this section, we study the effectiveness of \model ~as a sequence modeling framework for RL.
We compare our \model ~with DT variants, including the original DT \cite{chen2021decision}, Decision S4 (DS4) \cite{david2023decision}, and Decision ConvFormer (DC) \cite{kim2024decision}.
In particular, we adopt the experimental setup from the DC paper to ensure a fair comparison.
We evaluate the model on both continuous OpenAI Gym \cite{brockman2016openai} and discrete Atari \cite{bellemare2013arcade} control tasks.
Details of each domain can be found in the corresponding subsections below.
All the training for the model is conducted on a single A100 GPU.%
\footnote{%
    The code of our experiments is available at \url{\githubrepo}.
}

\subsection{OpenAI Gym}

\begin{table}[t]
    \centering
    \caption{
        The results for D4RL datasets.
        We report the expert-normalized returns, following \cite{fu2020d4rl}, averaged across five random seeds.
        }
    \begin{tabular}{lcccc}
        \toprule
        Dataset         & DT    & DS4   & DC    & \textbf{\model}             \\
        \midrule
        HalfCheetah-m   & 42.6  & 42.5  & 43.0  & 42.8\scriptsize{$\pm$0.08}  \\
        Hopper-m        & 68.4  & 54.2  & 92.5  & 83.5\scriptsize{$\pm$12.5}  \\
        Walker-m        & 75.5  & 78.0  & 79.2  & 78.2\scriptsize{$\pm$0.6}   \\
        \cmidrule(rl){1-5}
        HalfCheetah-m-r & 37.0  & 15.2  & 41.3  & 39.6\scriptsize{$\pm$0.1}   \\
        Hopper-m-r      & 85.6  & 49.6  & 94.2  & 82.6\scriptsize{$\pm$4.6}   \\
        Walker-m-r      & 71.2  & 69.0  & 76.6  & 70.9\scriptsize{$\pm$4.3}   \\
        \cmidrule(rl){1-5}
        HalfCheetah-m-e & 88.8  & 92.7  & 93.0  & 91.9\scriptsize{$\pm$0.6}   \\
        Hopper-m-e      & 109.6 & 110.8 & 110.4 & 111.1\scriptsize{$\pm$0.3}  \\
        Walker-m-e      & 109.3 & 105.7 & 109.6 & 108.3\scriptsize{$\pm$0.5}  \\
        \bottomrule
    \end{tabular}
    \label{tab:results-gym}
\end{table}

We consider here the domain of continuous control tasks from the D4RL benchmark \cite{fu2020d4rl},
which features several continuous locomotion tasks with dense rewards.
We conduct experiments in three environments: HalfCheetah, Hopper, and Walker.
For each environment, we examine three distinct datasets, each reflecting a different data quality level:
\begin{itemize}
    \item Medium (m): one million timesteps generated by a policy that achieves approximately one-third the score of an expert policy
    \item Medium-Replay (m-r): the replay buffer of an agent trained to match the performance of the Medium policy
    \item Medium-Expert (m-e): one million timesteps generated by the Medium policy added with one million timesteps generated by an expert policy
\end{itemize}

The overall results are shown in \tref{tab:results-gym}.
The scores are normalized so that a score of 100 represents the performance of an expert policy, following the methodology in \cite{fu2020d4rl} (see \aref{appendix:scores} for details).
The results for DT, DS4, and DC are quoted from the DC paper, Table 1 therein.
To ensure a fair comparison with \model, we set the hyperparameters identical to those reported in \cite{kim2024decision}.
The details of the experimental setups are provided in \aref{appendix:gym}.

\subsection{Atari}

\begin{table}[t]
    \centering
    \caption{
        The results for the 1\% DQN-replay Atari datasets.
        We report the gamer-normalized returns, following \cite{ye2021mastering}, averaged across three random seeds.
        }
    \begin{tabular}{lccc}
        \toprule
        Game      & DT                          & DC                          & \textbf{\model}              \\
        \midrule
        Breakout  & 242.4\scriptsize{$\pm$31.8} & 352.7\scriptsize{$\pm$44.7} & 239.2\scriptsize{$\pm$26.4}  \\
        Qbert     & 28.8\scriptsize{$\pm$10.3}  & 67.0\scriptsize{$\pm$14.7}  & 42.3\scriptsize{$\pm$8.5}    \\
        Pong      & 105.6\scriptsize{$\pm$2.9}  & 106.5\scriptsize{$\pm$2.0}  & 63.2\scriptsize{$\pm$102.1}  \\
        Seaquest  & 2.7\scriptsize{$\pm$0.7}    & 2.6\scriptsize{$\pm$0.3}    & 2.2\scriptsize{$\pm$0.03}    \\
        \bottomrule
    \end{tabular}
    \label{tab:results-atari}
\end{table}

The Atari domain is built upon a collection of classic video games \cite{mnih2013playing}.
In this domain, we have a discrete action space and face the challenge of long-term credit assignment.
This challenge arises from the delay between actions and their corresponding rewards.
For such types of data, the Mamba framework could potentially offer an advantage over prior SSMs, such as S4.
We consider four Atari games: Breakout, Qbert, Pong, and Seaquest, which are evaluated in \cite{agarwal2020optimistic}.
Following the setup in \cite{agarwal2020optimistic}, we train the model on 1\% of all samples in the DQN-replay dataset.
This equates to 500,000 transitions out of the 50 million collected by an online DQN agent during training \cite{mnih2015human}.

Table \ref{tab:results-atari} shows the results in the Atari domain.
We normalize the scores based on a professional gamer's score as per \cite{ye2021mastering}, with 100 representing a professional gamer and 0 a random policy.
For DT and DC, we quote the scores directly from Table 3 of the DC paper.
The training setups are identical to those reported in \cite{kim2024decision}.
For the details, see \aref{appendix:atari}.

\subsection{Ablation study}

To investigate the capability of the Mamba block in \model, we consider ablation studies on the channel-mixing layer and the context length $K$.

\begin{table}[ht]
    \centering
    \caption{
        Ablation study for the channel-mixing blocks in \model ~for D4RL datasets.
        We report the expert-normalized returns averaged across five random seeds.
        (1) Remove channel-mixing layer, and (2) Remove channel-mixing layer and double the total number of layers.
        }
    \begin{tabular}{lccc}
        \toprule
        Mamba layer    & HalfCheetah                 & Hopper                      & Walker                     \\
        \midrule
        Default        & 42.8\scriptsize{$\pm$0.08}  & 83.5\scriptsize{$\pm$12.5}  & 78.2\scriptsize{$\pm$0.6}  \\
        (1) RC         & 42.9\scriptsize{$\pm$0.01}  & 82.9\scriptsize{$\pm$5.8}   & 78.0\scriptsize{$\pm$2.0}  \\
        (2) RC \& $2L$ & 42.8\scriptsize{$\pm$0.06}  & 85.2\scriptsize{$\pm$14.9}  & 77.0\scriptsize{$\pm$1.1}  \\
        \bottomrule
    \end{tabular}
    \label{tab:ablation-gym}
\end{table}

We first study the contribution of the channel-mixing blocks of the Mamba layers.
As mentioned in the previous section, we included the channel-mixing layers in \model ~to make the experimental setups equivalent to the baselines.
Given that the channel-mixing layer is not a core component of the original Mamba model—the Mamba block itself integrates both token- and channel-mixing functions—we remove the channel-mixing blocks from all the Mamba layers.
We then train the whole network on the D4RL datasets from scratch in two configurations, keeping the rest unchanged:
\begin{enumerate}
    \item Remove the channel-mixing layer (RC)
    \item Remove the channel-mixing layer and double the total number of layers ($2L$)
\end{enumerate}

\tref{tab:ablation-gym} shows the results. We here consider only the Medium type datasets.
From \tref{tab:ablation-gym}, we observe that \model ~really demonstrates the comparable performance without the channel-mixing layers.
This suggests, as expected from the Mamba paper discussion, that the Mamba block suffices for RL sequence modeling tasks within this empirical scope.

We also examined the effect of context length $K$ on \model.
As discussed in the Mamba paper, the Mamba block generically exhibits robust performance for longer sequence lengths.
To explore the implications of context length $K$, we utilize datasets from the Atari domain, specifically Breakout and Qbert.
This choice was made because, in OpenAI Gym environments, context length is not expected to significantly affect performance due to their MDP formulation.

\begin{table}[ht]
    \centering
    \caption{
        Ablation study for the context length $K$ for \model ~in the Atari domain.
        We report the gamer-normalized returns averaged across three random seeds.
        }
    \begin{tabular}{ccc}
        \toprule
        $K$ & Breakout                    & Qbert                       \\
        \midrule
        10  & 231.6\scriptsize{$\pm$16.2} & 56.4\scriptsize{$\pm$16.6}  \\
        30  & 239.2\scriptsize{$\pm$26.4} & 42.3\scriptsize{$\pm$8.5}   \\
        40  & 295.9\scriptsize{$\pm$34.7} & 28.1\scriptsize{$\pm$4.5}   \\
        60  & 271.1\scriptsize{$\pm$70.8} & 15.7\scriptsize{$\pm$5.5}   \\
        \bottomrule
    \end{tabular}
    \label{tab:ablation-atari}
\end{table}

The results are shown in \tref{tab:ablation-atari}.
We set the context lengths $K=10$, $40$, and $60$, where $K=30$ is the default, and train the model from scratch.
The rest of the training configurations is taken exactly the same as in the previous subsection.
\tref{tab:ablation-atari} tells us that for Breakout the longer context lengths generically improve the performance.
The results for Qbert show that the longer context lengths significantly degrade the performance.
This may imply that in training the model for Qbert, the selection mechanism in the Mamba block does not work well for a longer context.
Rather, it might hinder generalization by removing indispensable tokens.

%% file: sec5.tex
\section{Conclusion and Discussion}

In this work, we examined the capabilities of the recently proposed Mamba for sequence modeling in RL.
We introduced Decision Mamba, which incorporates the Mamba block based on the selective SSM, into the DT-type neural network architecture.
Our empirical study shows that \model ~is competitive to existing DT-type models, suggesting an effectiveness of Mamba for RL tasks.
We hope this investigation offers insights into sequential decision-making, potentially paving the way for significant advancements in the field.

In this paper, we have not explored the efficiency perspectives of the model, one of the key contributions of Mamba enabled by a hardware-aware parallel algorithm \cite{gu2023mamba}.
In fact, merely applying the Mamba block to DT-type networks does not enhance efficiency due to numerous interactions between CPUs and GPUs for RL tasks considered in this paper, as seen in both training and inference phases.
Although we adhere to DT's experimental setups for fair comparison, it is desirable to reconsider the implementation to leverage Mamba's advantages efficiently.

Another limitation of this study is the absence of a hyperparameter search and an analysis of how to use the Mamba block more effectively to reflect the data structure of RL tasks.
Unlike the typical one-dimensional sequential data found in natural language processing, for instance, a trajectory from an MDP features a unique structure comprising states, actions, and rewards sequentially.
Furthermore, in conventional RL problems, immediate states and rewards are crucial for decision-making due to the inherent nature of MDPs.
To fully explore Mamba's potential in RL problems, one approach could be to adapt the network architecture to better suit the RL data structure or preprocess trajectory datasets into a format more compatible with Mamba.
Another avenue to explore is applying Mamba to non-Markov decision processes, notorious for their complexity due to long-range interactions within trajectories.
We leave these aspects, including parameter tuning, for future work.
A comprehensive study with an exploration of improved Transformer-type network architectures for RL will be released elsewhere.

%% file: appendix.tex
\section{Implementation Details of \model}
\label{appendix:implementation}

\subsection{Code of the Mamba layer}
\label{appendix:code}

We implement \model ~using the official DT code%
\footnote{%
    \url{https://github.com/kzl/decision-transformer}.
    We also employ the DC code for some considerations, which is available as a supplementary material at \url{https://openreview.net/forum?id=af2c8EaKl8}.
}
and incorporate the Mamba module.
The pseudo-code of the Mamba layer is shown in Algorithm~\ref{alg:pseudocode} below.
The full set of code of our experiments is available at \url{\githubrepo}.

\begin{algorithm}
\caption{Pseudo-code of the Mamba layer, PyTorch-like code.}
\label{alg:pseudocode}
\begin{lstlisting}[language=Python]
from mamba_ssm import Mamba

class Block(nn.Module):
    def __init__(self, config):
        super().__init__()
        self.ln1 = nn.LayerNorm(config.n_embd)
        self.mamba = Mamba(config.n_embd)

        self.ln2 = nn.LayerNorm(config.n_embd)
        self.mlp_channels = nn.Sequential(
            nn.Linear(config.n_embd, 4 * config.n_embd),
            nn.GELU(),
            nn.Linear(4 * config.n_embd, config.n_embd),
            nn.Dropout(config.resid_pdrop),
        )

    def forward(self, x):
        x = x + self.mamba(self.ln1(x))
        x = x + self.mlp_channels(self.ln2(x))
        return x
\end{lstlisting}
\end{algorithm}

\subsection{OpenAI Gym}
\label{appendix:gym}

For the training on the OpenAI Gym tasks, the hyperparameters are adopted from \cite{chen2021decision} and \cite{kim2024decision}.
The common hyperparameters are shown in \tref{tab:hyper-gym}.

\begin{table}[ht]
    \centering
    \caption{Hyperparameters of \model ~on the D4RL datasets.}
    \label{tab:hyper-gym}
    \begin{tabular}{lc}
        \toprule
        Hyperparameter          & Value         \\
        \midrule
        Number of layers        & 3             \\
        Batch size              & 64            \\
        Context length $K$      & 20            \\
        Return-to-go conditioning & 6000 HalfCheetah \\
                                & 3600 Hopper   \\
                                & 5000 Walker   \\
        Dropout                 & 0.1           \\
        Nonlinearity function   & $\gelu$       \\
        Grad norm clip          & 0.25          \\
        Weight decay            & $10^{-4}$     \\
        Learning rate decay     & Linear warmup \\
        Total number of updates & $10^{5}$      \\
        \bottomrule
    \end{tabular}
\end{table}

Some of the hyperparameters are adapted for each dataset as in \cite{kim2024decision}:
\begin{itemize}
    \item Embedding dimension: 256 for Hopper-m, Hopper-m-r. 128 for other environments
    \item Learning rate: $10^{-4}$ for Hopper-m, Hopper-m-r, Walker-m. $10^{-3}$ for other environments
\end{itemize}

\subsection{Atari}
\label{appendix:atari}

Similarly to the OpenAI Gym tasks, the hyperparameters for the Atari domain follow those from \cite{kim2024decision}, as listed in \tref{tab:hyper-atari}.

\begin{table}[ht]
    \small
    \centering
    \caption{Hyperparameters of \model ~on the Atari games.}
    \label{tab:hyper-atari}
    \begin{tabular}{lc}
        \toprule
        Hyperparameter              & Value \\
        \midrule
        Number of layers            & 6     \\
        Embedding dimension         & 128   \\
        Batch size                  & 256   \\
        Context length $K$          & 30 Breakout, Qbert, Seaquest  \\
                                    & 50 Pong        \\
        Return-to-go conditioning   & 90 Breakout    \\
                                    & 14000 Qbert    \\
                                    & 20 Pong        \\
                                    & 1150 Seaquest  \\
        Nonlinearity                & $\operatorname{ReLU}$ Encoder   \\
                                    & $\gelu$ Otherwise               \\
        Max epochs                  & 10                              \\
        Dropout                     & 0.1                             \\
        Learning rate               & $6 \times 10^{-4}$              \\
        Adam betas                  & (0.9, 0.95)                     \\
        Grad norm clip              & 1.0                             \\
        Weight decay                & 0.1                             \\
        Learning rate decay         & Linear warmup and cosine decay  \\
        Warmup tokens               & $512 \times 20$                 \\
        Final tokens                & $2 \times 500000 \times K$      \\
        \bottomrule
    \end{tabular}
\end{table}

\section{Task Scores and Normalization}
\label{appendix:scores}

In the main text, we report the normalized scores for each domain calculated by the following rule with baseline random and expert scores:
\begin{equation}
    \text{score}_{\text{normalized}} := 100 \times \frac{\text{score}_{\text{raw}} - \text{score}_{\text{random}}}{\text{score}_{\text{expert}} - \text{score}_{\text{random}}}.
\end{equation}

For OpenAI Gym tasks, we employ the protocol from \cite{fu2020d4rl}.
For Atari games, we have raw and baseline scores as in \tref{tab:atari_main_raw}, which are used for normalization in \cite{ye2021mastering}, Table 1 therein.
The column of \model ~in \tref{tab:atari_main_raw} shows the raw scores corresponding to \tref{tab:results-atari} in the main text.

\begin{table}[ht]
    \centering
    \caption{
    The mean and variance of raw scores for the 1\% DQN-replay Atari datasets across three seeds.
    ``Random'' and ``Expert'' indicate the Atari baseline scores used for normalization.
    }
    \label{tab:atari_main_raw}
    \begin{tabular}{lccc}
        \toprule
        Game     & \model                         & Random  & Expert   \\  
        \midrule
        Breakout & 70.6\scriptsize{$\pm$9.3}      & 1.7     & 30.5     \\
        Qbert    & 5780.0\scriptsize{$\pm$1295.2} & 163.9   & 13455.0  \\
        Pong     & 1.6\scriptsize{$\pm$15.3}      & $-$20.7 & 14.6     \\
        Seaquest & 1006.0\scriptsize{$\pm$57.7}   & 68.4    & 42054.7  \\
        \bottomrule
    \end{tabular}
\end{table}